# Clinical Text Generation through Leveraging Medical Concept and Relations

**Wangjin Lee[†1], Hyeryun Park[†1], Jooyoung Yoon[1], Kyeongmo Kim[1], and Jinwook Choi[*1,2]**

[1]Interdisciplinary Program in Bioengineering, Graduate School, Seoul National University
[2]Department of Biomedical Engineering, College of Medicine, Seoul National University
{jinsamdol, Helena.park, joo0yoon, medinfoman,jinchoi}@snu.ac.kr

## Abstract

With a neural sequence generation model, this study aims to develop a method of writing the patient clinical texts given a brief medical history. As a proof-of-a-concept, we have demonstrated that it can be workable to use medical concept embedding in clinical text generation. Our model was based on the Sequence-to-Sequence architecture and trained with a large set of de-identified clinical text data. The quantitative result shows that our concept embedding method decreased the perplexity of the baseline architecture. Also, we discuss the analyzed results from a human evaluation performed by medical doctors.

## 1 Introduction

Clinical natural language processing is one of the interesting research field associated with medical artificial intelligence (AI). Compared to research in the general domain, researchers in the medical domain are relatively more difficult to acquire data for their study because most clinical documents are likely to contain sensitive information, making the data unapproachable.

Although data management techniques such as de-identification and anonymization can be good choices for publishing the clinical documents, our choice in this study is to use a language generation model to make our own clinical AI writer. The clinical AI writer can freely produce the designed output by taking the human's input into its account. Thus, using our system, humans only write the general description or brief medical history of a patient as the input set for the generation model and the model generates appropriately entailed clinical stories for the corresponding input (Fig 1).

In the medical domain, clinical text generation has long been receiving attention as a research topic. A recent study demonstrated the Chinese medical record generation (Guan et al., 2019) based on generative adversarial nets (GAN) (Goodfellow et al., 2014). Though the GAN is known to generate more readable texts than other models, it is known to be unstable at the training.

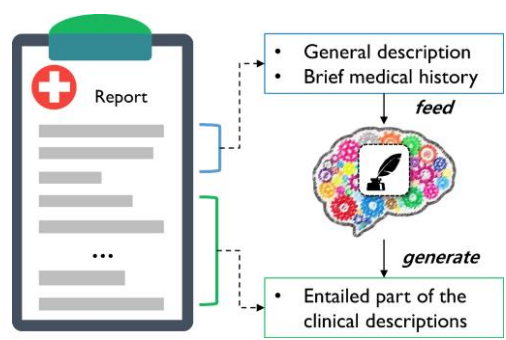

Figure 1. Generation model producing clinical documents given general description of a patient

One of deep learning approaches is using language models which are probabilistic models, predicting the next word of a given sequence. Encoder-decoder architecture (Cho et al., 2014) consists of 2 recurrent neural networks(RNNs). The Sequence-to-Sequence (Seq2Seq) (Sutskever et al., 2014) with Attention mechanism (Bahdanau et al., 2015) consists of 2 LSTMs based on encoder-decoder structure. The encoder provides input sequence as a context vector, the decoder decodes this vector to predict the next word and attention allows to attend different parts of the input sequence. The transformer model (Vaswani et al., 2017) is based on attention mechanisms, which captures global dependencies between input and output sequences. Due to the stability and the model's simplicity, we tested the Seq2Seq for the generation of clinical texts in this study. Given several sentences from the beginning of the clinical notes, we trained the model to generate the rest part of it.

Knowledge base are utilized to generate more semantically correct, coherent and informative text. Knowledge graph consists of triplets (entity1, relation, entity2) representing relational facts, for example WordNet (Miller, 1995) and Freebase(Bollacker et al., 2008). As clinicians have a unique way of describing disease conditions, we designed Seq2Seq enforced with medical concept embedding. In order to embed medical knowledge that can be seen as triplets, medical concept codes were selected as the auxiliary input of the Seq2Seq. Thus we leveraged the medical concept unique identifiers (CUIs) from the UMLS (unified medical language system) (Lindberg et al., 1993) for the domain knowledge.

According to Dusek's work (Dušek and Jurčíček, 2016), the auxiliary context in coded form may be either appended to the original word sequence or separately encoded in parallel. Thus, we tested both Seq2Seq encoder structures to appropriately provide the context from the CUIs to the decoder RNN. For the proof of concept, we demonstrate an approach of the clinical text generation handling clinical concepts in a medical thesaurus and embedding the concepts from the hierarchical tree.

## 2 Clinical text and medical thesaurus

Two databases were used in this study: MIMIC-III (Medical Information Mart for Intensive Care) and UMLS. The MIMIC-III is a large-scale health-related database (Johnson et al., 2016). It contains clinical notes having narrative descriptions of patients' previous history and current progress.

The UMLS is a thesaurus consisting of multiple medical terminology systems. A CUI is a concept identifier in the system. For the naming convention, a CUI consists of eight characters: the letter 'C' and seven numbers following the letter. The CUI in the UMLS represents the highly specific concept of a medical entity. Because a concept has multiple different names, a single CUI can be representative of multiple medical names.

We embedded the CUIs by utilizing the semantic relationship between medical concepts in the UMLS. In a simplified abstraction, we may say the relationships can be written in the form of a triplet having elements as two CUIs and one relation label. For instance, 'Breast carcinoma' (C0678222) has child-parent (*PAR*) relationship with both 'Malignant neoplasms' (C0006142) and 'Breast diseases' (C0006145), then, these relationships can be written as (C0678222, C0006142, *PAR*) and (C0678222, C0006145, *PAR*). We mainly used this triplet for our CUI embedding.

### 2.1 Data preprocessing

The de-identified clinical notes are preprocessed in multiple steps. SpaCy (Honnibal and Montani, 2017) was used in order to extract sentences from the documents. The sentence boundary detection module of the SpaCy was applied to the text. With the syntactic parsing module, we selected narrative sentences satisfying sentence structure that contains either subject or object element.

After that, we tried to choose sentences that had informative contents. At the first step, CUIs were extracted from sentences. This step is mapping medical knowledge that is latent throughout a sentence into codes. MetaMap (Aronson, 2001) was utilized for identifying UMLS concepts from the text. At this process, sentences having at least one UMLS CUI were selected. Then, we extracted informative sentences in a document based on the Shannon entropy.

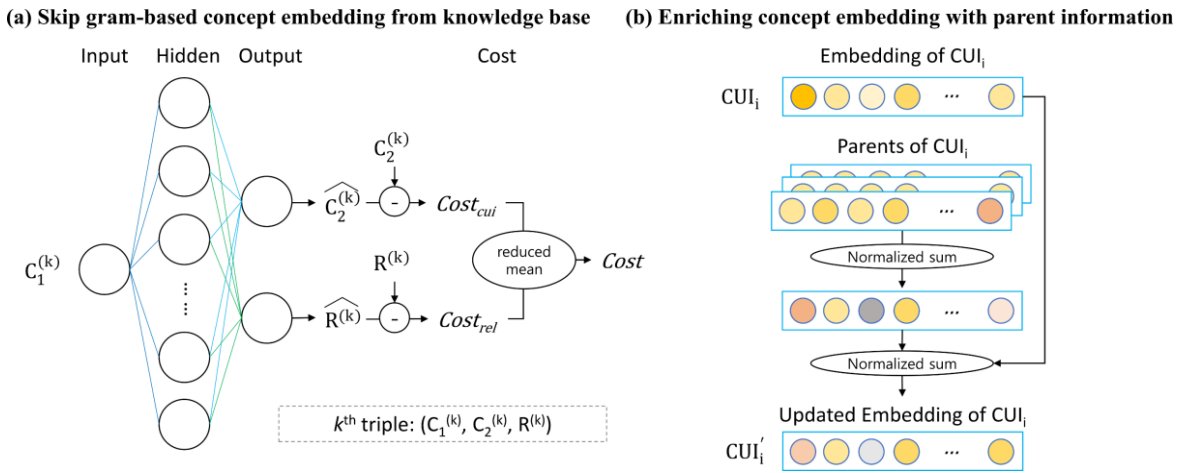

Figure 2. Graphical abstraction of the concept embedding.

## 3 Hierarchical concept embedding

Our clinical text generation pipeline is based on the Seq2Seq. At the encoding phase, we used medical concept information that is a set of CUIs mapped to text to make context beyond the word itself. To use the concept, it was necessary to make embedding of the concepts. Figure 2 shows the abstraction of our concept embedding approach.

### 3.1 Concept embedding

The neural architecture suggested in this study consists of one input, one hidden, and one output layer. Given one concept as an input, it is trained to bring another CUI that has any relationship with the input and jointly bring the corresponding relation label. The weight vector between the input and the hidden layer is used as the embedding of the CUIs.

Figure 2-a shows the concept embedding architecture. The $k^{th}$ triplet can be ($C_1^{(k)}$, $C_2^{(k)}$, $R^{(k)}$) and each notation is for the first CUI, the second CUI, and the relation label, respectively. The set of the $C_1$ is obtained from the training data and the $C_2$s are ones having relationships with the $C_1$ in UMLS. We would note that our procedure is motivated by the Skip-gram (Mikolov et al., 2013): instead of recalling neighbor words in the Skip-gram, our approach pursues the objective of jointly recalling the related CUI and the relation label.

Because our architecture cooperatively produces two outputs, the final cost is the average of the intermediate costs calculated with cross-entropy using one-hot encoding.

### 3.2 Enriching the concept embedding

The implementation of the concept embedding is based on the idea that the vector representation of a concept should be closer to others if they share more entities. Our idea enriching the concept embedding is to make the vector of a specific concept close with the vectors of its parents in the hierarchy tree.

This idea is motivated by FastText (Bojanowski et al., 2017). The skip-gram model ignores the morphology of words, but the FastText makes a word vector as the sum of sub-word vectors. Instead of summing vectors of sub-words, our hierarchical concept embedding algorithm sums up embedding vectors of parent concepts and the vector of the concept itself. We initially have a vector $c_i$ for the $i^{th}$ CUI from the skip-gram based concept embedding. Then, $c'_i$, the updated vector of the $c_i$ is calculated as following (Eq. 1).

$$c'_i = \frac{1}{2}\left(c_i + \frac{1}{|Par_i|}\sum_{p \in Par_i} c_p\right) \quad (1),$$

where $Par_i$ is the parent set of $c_i$ (Figure 2-b). Embedding of unknown concepts are dynamically calculated by $l2$-normalization on the sum of vectors which concept that has a common prefix (length=7) and also the vectors which concept has occurred once.

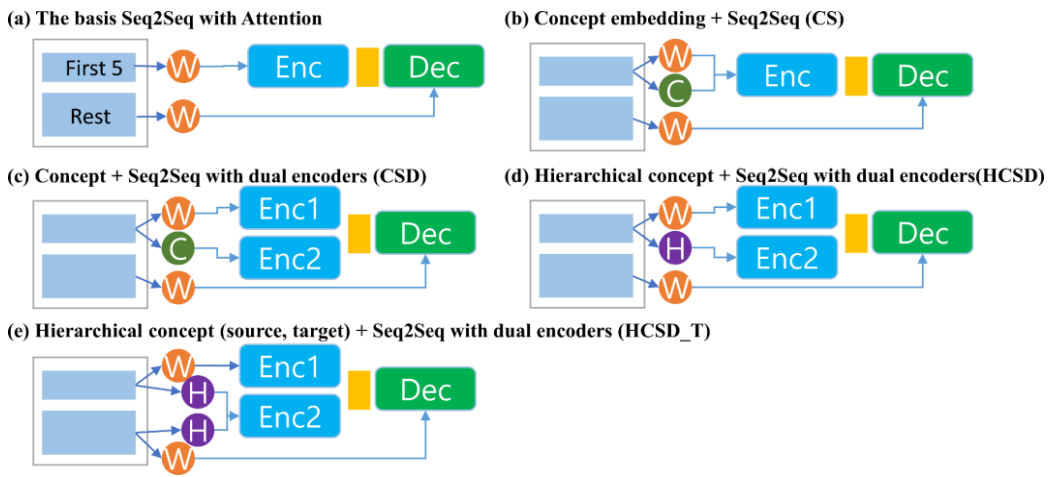

Figure 3. Box plots of the experts' ratings on the generated clinical texts and the human's text.

## 4 Model design

Figure 3 shows 5 model structure we designed. The basis of our model is Seq2Seq with Attention mechanism. The input is a general description or brief medical history of the patient in the document, and the output is the rest part of the document. In this study, it was hard to identify the part on the general descrition, thus we assume the first five sentences of a clinical note as the input: according to our data observation, the beginning part of a clinical text gives a more general description of the patient than the latter part. The input is expected to provide the decoder a summarized representation of the patient's general description as a context vector. The decoder output was the rest part of the clinical note during the training time.

### 4.1 Incorporating medical concepts

We fed word sequence as well as a CUI sequence extracted from the source sentences with MetaMap into the encoder. The CUI is expected to lead the decoder to predict word that are semantically closer to the input in terms of domain knowledge.

We tested two types of encoder structures demonstrated in (Dušek and Jurčíček, 2016) to provide auxiliary information. The first one is to concatenate the CUI sequence at the end of the word sequence in a single encoder (CS; Concept+Seq2Seq). The second structure is a Seq2Seq having dual encoders (CSD; Concept+SeqSeq with dual encoders). One encoder of the CSD makes a context vector from the first five sentences and the other makes a context vector from the CUI sequence. The encoded results from each encoder in the CSD are concatenated before going inward the decoder.

## 5 Evaluation settings and results

Our task was to generate clinical descriptions from the first introductory part of the full description. Thus, the source was the first five sentences and the target was the rest sentences. The embedding vectors were trained with 100,000 training set. Out-of-vocabulary tokens are prevalent in clinical notes. The reason we used FastText for word embedding is to handle these out-of-vocabulary tokens. The vocabulary sizes were 46,975 and 49,758 for source and target respectively. For the concept embedding, the CUIs came from the training data, and the number of relationships extracted from the UMLS was 61,299,702 consisting of 50,942 CUIs for the $C_1$, 1,005,865 CUIs for the $C_2$, and 672 relation labels.

The language models were trained with 35,000 notes selected from the training set and validated with 8,603 notes from another set. The number of test set was 8,578. We evaluated five settings: they were named according to the method of the concept embedding. The baseline was the Seq2Seq with Attention mechanism. The baseline was trained without concepts. The others were given the sentence as well as the CUI information with different model structures and different CUI embedding methods. The second and the third models (CS, CSD) were tested in order to compare the encoder structure (single vs. dual). The CUI embedding method for the models was the skip-gram based concept embedding (§3.1). The last two models (HCSD and HCSD_T) used the hierarchical concept embedding (§3.2) in the dual encoder structure, and the last one utilized CUIs of both source and target text. The RNN unit was three-layer Bi-LSTM and the RNN size was 400.

Table 1 shows the models' perplexity. Because the text generation task is an open-ended problem, a common evaluation method for this task is the perplexity. The results indicate that some models using the concept showed lower complexity than Seq2Seq trained without domain knowledge. The model CS using concepts in single encoder reduced the test perplexity by 0.432 than the baseline. The HCSD using the hierarchical concept embedding in dual encoder showed subtle improvement.

Table 1. Perplexities of the clinical text generation models on MIMIC-III.

| Model | Valid perplexity | Test perplexity |
|---|---|---|
| Seq2Seq | 3.423 | 3.800 |
| CS | **3.360** | **3.368** |
| CSD | 3.822 | 4.195 |
| HCSD | 3.702 | 3.764 |
| HCSD_T | 3.830 | 4.197 |

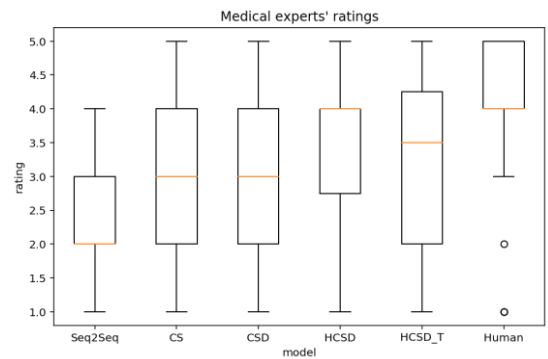

Figure 4. Box plots of the experts' ratings on the generated clinical texts and the human's text.

**Human evaluation with medical experts:** Four medical doctors evaluated the clinical notes in terms of quality. The expert group consists of four medical doctors with an average of 10 years of experience. Duplicates in the generated texts were removed for this human evaluation. The questionnaire consists of two chapters: the first part asks evaluators to rate the text generated for the input sentence looks like a human text, and the second part asks them to identify the human-written paragraph in the series of texts. Each part consists of ten questions (the full document is in the Appendix).

In the first chapter, the evaluators were given paragraphs written by five models and human-written paragraph. Not knowing which paragraph is written by human, they rated on a 5-point scale(1: very awkward, 5:very plausible). This assessment focused on the logical appropriateness, accuracy in terms of medical knowledge, and how natural writing is about the input. To prevent bias, we did not provide any information about the models as well as the fact that the paragraph contains human writing. Figure 4 shows boxplots summarizing the ratings from the evaluators. The plot shows that the models using the concepts were rated higher than the basic model, and the best model (HCSD) achieved a median value that is equal to the median value for the human's writings (median rate=4). The models using hierarchical concept embedding also seem to achieve better performance than the models not using the hierarchy. The shape of the box for the HCSD indicates that the rating scores were more closely distributed to the median than other models. An interesting point was that the humans' writings were sometimes rated low (outlier circles under the bar): this phenomenon is presumably as human's clinical writings sometimes do not seem to be formal and contain unaccustomed expressions even for experts, however, this needs further investigation.

In the second part, the evaluators were given a set of three separate texts produced by either the generation models or human writer, and they were asked to identify a paragraph which was written by a human. We measured error rate of the evaluators who identified human writing, with an average error rate of 65% (Table 2). Also, the evaluators reported they struggled to perform this task (2.63 is close to the highest level of difficulty). For the purpose of comparison, the error rate from four non-experts are presented (who are graduate students in Biomedical Engineering). This result indicate that the clinical text generation models can produce virtual text similar to real text written by human.

Some evaluators reported the reason for their answers. Significant evidences for recognizing artificial writers were that repetitive sentence structure and ridiculous expression (e.g., '*the left ventricle is not clearly seen, but the left ventricle is not clearly seen.*'). Also, they thought a text was written by a human writer when the text contains causal relationships or when the description was in chronological order. This observation provides

Table 2. Error rates (%) for the identification of human's writing and the difficulty level reported by evaluators (1: easy, 3: confused.)

|  | **Expert** | **Non-expert** |
|---|---|---|
| Error rate (average) | 65 | 75 |
| Level of difficulty (1-3) | 2.63 | 2.58 |

clues for planning the direction of further study.

## 6 Conclusion

In this paper, we demonstrate a clinical text generation method based on the Seq2Seq model. Because this is a preliminary study, the current model have more rooms to be improved. Our approach feeding concept information in the sequence generation research of clinical domain would improve the quality of the generated texts. We plan to study the concept embedding method more in-depth with state-of-the-art generative models for the same task in the future.

## References

Alan R Aronson. 2001. Effective Mapping of Biomedical Text to the UMLS Metathesaurus : The MetaMap Program. :17–21.

Dzmitry Bahdanau, Kyunghyun Cho, and Yoshua Bengio. 2015. Neural Machine Translation by Jointly Learning to Align and Translate. In *ICLR*, pages 1–15.

Piotr Bojanowski, Edouard Grave, Armand Joulin, and Tomas Mikolov. 2017. Enriching Word Vectors with Subword Information. In *Transactions of the Association for Computational Linguistics*, pages 135–146.

Ondřej Dušek and Filip Jurčíček. 2016. A Context-aware Natural Language Generator for Dialogue Systems. In *Proceedings of the SIGDIAL 2016 Conference*, pages 185–190.

Ian Goodfellow, Jean Pouget-Abadie, Mehdi Mirza, Bing Xu, David Warde-Farley, Sherjil Ozair, Aaron Courville, and Yoshua Bengio. 2014. Generative Adversarial Nets. In *NIPS2*.

Jiaqi Guan, Runzhe Li, Sheng Yu, and Xuegong Zhang. 2019. Generation of Synthetic Electronic Medical Record Text. In *Proceedings - 2018 IEEE International Conference on Bioinformatics and Biomedicine, BIBM 2018*, pages 374–380.

Matthew Honnibal and Ines Montani. 2017. spaCy 2: Natural language understanding with Bloom embeddings, convolutional neural networks and incremental parsing. *To appear*.

Alistair E W Johnson, Tom J Pollard, Lu Shen, Li-Wei H Lehman, Mengling Feng, Mohammad Ghassemi, Benjamin Moody, Peter Szolovits, Leo Anthony Celi, and Roger G Mark. 2016. MIMIC-III, a freely accessible critical care database. *Scientific data*, 3:160035.

Patty Kostkova, Helen Brewer, Simon de Lusignan, Edward Fottrell, Ben Goldacre, Graham Hart, Phil Koczan, Peter Knight, Corinne Marsolier, Rachel A. McKendry, Emma Ross, Angela Sasse, Ralph Sullivan, Sarah Chaytor, Olivia Stevenson, Raquel Velho, and John Tooke. 2016. Who Owns the Data? Open Data for Healthcare. *Frontiers in Public Health*, 4(February).

Donald A Lindberg, Betsy L Humphreys, and Alexa T McCray. 1993. The Unified Medical Language System. *Methods of information in medicine*, 32(4):281–291.

Tomas Mikolov, Kai Chen, Greg Corrado, and Jeffrey Dean. 2013. Efficient Estimation of Word Representations in Vector Space. In *arXiv preprint arXiv:1301.3781*, pages 1–12.

Ilya Sutskever, Oriol Vinyals, and Quoc V. Le. 2014. Sequence to Sequence Learning with Neural Networks. In *NIPS 2014*, pages 1–9.

# 7 Supplementary Material (appendix file)

## 7.1 Model training setting

## 7.2 Human evaluation questionnaire

# Clinical Text Generation through Leveraging Medical Concepts and Relations

**Supplementary materials**

## 7.1 Model training setting and model designs

Word embedding dimension: 80

Word embedding method: FastText (in the genism library for Python)

Concept embedding dimension: 80

RNN size: 400

RNN model: Bidirectional LSTM, three layers

Attention: Bahdanau attention

Generalization: Dropout (keep_prob=0.9)

Epoch: 10~20

Beam search width: 1

Gradient clipping: [-1, +1]

## 7.2 Human evaluation questionnaire (following pages)

# Pseudo-clinical text generation evaluation

    This document is for the evaluation of English clinical text generation models. This consists of two chapters and each chapter has 10 questions. Please follow instructions and fill the answers.

〈Authors – anonymized〉
*The data used in the model training was a part of MIMIC-III dataset.

# Evaluator information

1. Occupation

    ○ medical student

    ○ intern

    ○ professor (research professor, fellow)

2. Department

    (                    )

3. Career

    (                        ) year

4. Sex

    ○ Male

    ○ Female

5. Age

    (                    )

# [Chapter 1: evaluate the appropriateness of the entailment]

<Instruction>

Given input is the first beginning part of a patient's clinical text. Following paragraphs are entailed texts to the input and they are independent each other.

Evaluators will measure each paragraph in the terms of following sense in 5-point Likert scale. (1: very awkward, 5: strongly likely)

- Does the paragraph make sense in the terms of clinical commonsense?
- Does the paragraph make sense in the terms of context?

☆ The paragraph number is not related to the model's id.

☆ Each paragraph has no relationship each other. They are following the input text individually.

☆ [deidt] label in the sentence is replacement tag for de-identification.

# Ch1. Appropriateness of entailment

Rate each output like this example

| Input |
|---|
| Portable ap radiograph of the chest was performed . |
| Comparison is made with prior radiograph dated [deidt] . |
| The et tube has its tip approximately 12 mm from the carina and is unchanged from the prior examination . |
| There is stable position of the left upper quadrant splenic drain and gastroesophageal stent . |

| No. | Output | Rating (Awkward ↔ appropriate) |
|---|---|---|
| 1 | There is a right ij central venous catheter with the tip in the right atrium . | (1) ○  (2) ●  (3) ○  (4) ○  (5) ○ |
| 2 | The cardiomediastinal silhouette is stable .<br>There is a left internal jugular catheter with tip in the right brachiocephalic vein .<br>The lungs are clear with no consolidation or pulmonary edema .<br>The osseous structures are unchanged . | (1) ○  (2) ○  (3) ●  (4) ○  (5) ○ |
| 3 | There is a right ij central venous catheter with the tip in the mid svc .<br>The cardiomediastinal silhouette is stable .<br>There is a left retrocardiac opacity which is unchanged .<br>The right lung is grossly clear . | (1) ○  (2) ●  (3) ○  (4) ○  (5) ○ |
| 4 | There is a small amount of free air in the abdomen .<br>There is no evidence of pneumothorax . | (1) ●  (2) ○  (3) ○  (4) ○  (5) ○ |
| 5 | There is a right internal jugular catheter with its tip in the right brachiocephalic vein .<br>There is a left subclavian line with its tip in the right atrium . | (1) ○  (2) ○  (3) ●  (4) ○  (5) ○ |
| 6 | There is stable appearance to the left basal atelectasis and there is slight interval increase in the right basal effusion and atelectasis .<br>The et tube has its tip approximately 12 mm from the carina and is unchanged from the prior examination , this needs to be repositioned .<br>Findings discussed with [deidt] | (1) ○  (2) ○  (3) ○  (4) ○  (5) ● |

Comment

# Ch1. Appropriateness of entailment

## Question 1

| Input |
|---|
| Reason : please assess for residual tumor |
| Admitting diagnosis : intracranial mass contrast : magnevist amt : 17 _ |
| T1 sagittal axial and flair t2 susceptibility and diffusion axial images were obtained before gadolinium . |
| Comparison was made with the mri of [deidt] . |

| No. | Output | Rating 매우 이상함◄►매우 적절함 |
|---|---|---|
| 1 | Comparison was made with the previous mri of [deidt] .<br>There is mild enhancement in the resection cavity indicative of residual tumor .<br>There is no evidence of abnormal enhancement seen .<br>There is mild prominence of temporal [deidt] which could be secondary to radiation changes .<br>The ventricles and sulci are normal in size and configuration . | ○ ○ ○ ○ ○<br>(1) (2) (3) (4) (5) |
| 2 | There is a right frontal approach ventricular catheter with a ventricular drain in place .<br>The previously seen right frontal ventricular catheter is in position .<br>The previously seen left frontal lesion is not seen on the current study .<br>The previously seen left frontal lobe mass has decreased in size since the study of [deidt] .<br>The previously seen enhancing lesion in the right parietal lobe has almost completely resolved with no significant enhancement identified . | ○ ○ ○ ○ ○<br>(1) (2) (3) (4) (5) |
| 3 | There is a small amount of fluid level in the right maxillary sinus .<br>There is also evidence of a small area of increased signal in the left side of the sphenoid sinus .<br>This is unclear from the patient 's void .<br>There is a large amount of inspissated secretions in the nasopharynx . | ○ ○ ○ ○ ○<br>(1) (2) (3) (4) (5) |
| 4 | There is mild prominence of ventricles and sulci .<br>There is no evidence of abnormal signal seen within the left heart or left heart . | ○ ○ ○ ○ ○<br>(1) (2) (3) (4) (5) |
| 5 | There are blood products seen in the surgical area .<br>Again , extensive edema is identified in the right cerebral hemisphere with mass effect on the right lateral ventricle and midline shift with mild periventricular edema along the left lateral ventricle .<br>Following gadolinium administration , some residual enhancement is identified posterior to the blood products in the right temporal region .<br>The stricture diffusion seen in the region appears to be secondary to blood products . | ○ ○ ○ ○ ○<br>(1) (2) (3) (4) (5) |
| | Mass effect again identified on the right lateral ventricle has slightly decreased . Midline shift uncal herniation and subfalcine herniation are noted with mild dilatation of the left lateral ventricle with periventricular edema as before . | |
| 6 | The right atrium is mildly dilated with a normal cavity motion . The right ventricular cavity size is normal with mild to severe left ventricular systolic function with global left ventricular systolic function with mild to moderate global hypokinesis . The right ventricular cavity size is normal with mild global free wall fluid disease . from the site of the patient 's clinical exam , the patient was agitated with gadolinium ( as compared to the study of [deidt] ) . Reason : please assess for residual tumor burden Admitting diagnosis : intracranial mass contrast : magnevist amt : 16 _ | ○ ○ ○ ○ ○ <br> (1) (2) (3) (4) (5) |

Comment

# Ch1. Appropriateness of entailment

## Question 2

| Input |
|---|
| This is a 65-year-old male who had experienced a week 's worth of crescendo angina , chest pain and shortness of breath . |
| His past cardiac history includes two weeks of angina in [deidt] that ultimately ended up in cardiac catheterization in [deidt] , leading to diagnosis of a complex left anterior descending lesion . |
| Subsequent to this , he had a recurrence of his anginal symptoms in [deidt] , therefore prompting another cardiac catheterization which showed a re-narrowing of the similar left anterior descending lesion and consequently underwent percutaneous transluminal coronary angioplasty brachytherapy , and cutting balloon . |
| After completing that , the patient was chest pain-free until approximately one week prior to this admission , when he began with the crescendo anginal symptoms , chest pain and shortness of breath that prompted him to be admitted on [deidt] , where he underwent cardiac catheterization showing a right dominant circulation system with a 70 % left anterior descending lesion with 90 % distal disease past the site of the stent . |
| The left circumflex had 30 % disease , and 90 % ostial lesion at the first run-off from the circumflex . |

| No. | Output | Rating 매우 이상함◀▶매우 적절함 |
|---|---|---|
| 1 | His preoperative laboratories were notable for a hematocrit of 26 , bun and creatinine of 19 and .9 .<br>: he was admitted to the c-med service , where he was ultimately referred to the cardiac surgical care unit with dr .<br>[deidt] revealed that he necessitated coronary artery bypass graft , and the patient subsequently went to the operating room on [deidt] .<br>In the operating room , doctor. [deidt] performed a coronary artery bypass graft x 4 , with left internal mammary artery to left anterior descending , saphenous vein graft to diagonal , saphenous vein graft to obtuse marginal i , sequentially to the obtuse marginal iii .<br>He had a right radial a-line , a right internal jugular swan-ganz catheter , two ventricular and two atrial leads , well as two mediastinal and one left pleural chest tube .<br>He was brought to the cardiac surgical recovery unit , where his drips were weaned to amiodarone .5 , milrinone .6 , and neo-synephrine .5 to maintain cardiovascular parameters of 83 and sinus , blood pressure of 99/40 after two units transfused , well as shovel mask at 100 % for the arterial blood gas of 7.34/46/113/26/-1 .<br>His bun and creatinine were 19 and .7 on day number two , and he was continuing on oral amiodarone .<br>He was given a physical therapy and rehabilitation screening , which ultimately | ○ (1) ○ (2) ○ (3) ○ (4) ○ (5) |
| | revealed that he was ambulator and independent enough to warrant discharge to home . His hematocrit was 24 , his bun and creatinine were 20 and .9 respectively on postoperative day three . His vital signs were 99.2 for a t-max , 75 and sinus rhythm , 100/59 for blood pressure . | |
| 2 | The patient was taken to the operating room where he underwent coronary artery bypass grafting to the left anterior descending coronary artery . The patient was taken to extracorporeal room on [deidt] for a redo one of her sternum , and a chest tube was placed . On [deidt] , the patient was transferred to the floor , and on [deidt] , the patient was taken to the operating room where on postoperative day number five . The patient was taken to extracorporeal operating on [deidt] where a chest tube was placed on [deidt] , and the patient was taken off of the hospital . The patient was taken to extracorporeal operating on [deidt] where a followup chest x-ray showed a small amount of residual blood in the left upper pole . The patient was taken to extracorporeal room and was taken to extracorporeal room on [deidt] . | ○ ○ ○ ○ ○<br>(1) (2) (3) (4) (5) |
| 3 | The patient was admitted to the hospital on [deidt] where he underwent coronary artery bypass grafting times four with a lima off the pda . He tolerated the procedure well and was transferred to the cardiothoracic intensive care unit in stable condition . On postoperative day number five , the patient was transferred to the floor in stable condition . The patient was tolerating a regular diet , was discharged home with instructions for discharge . The patient was discharged home with roxicet elixir , and his diet was slowly advanced in the hospital . The patient was discharged home with vna services and follow-up followup with doctor. [deidt] in clinic . Please call your doctor or return to the emergency department if you have nausea chills , shortness of breath , chest pain , shortness of breath nausea vomiting , or any other symptoms . Please resume your blood pressure in the absolute period of on discharge . If you have a lot of blood sugars , you are more than night aspirin plavix , , coumadin , and for coumadin . | ○ ○ ○ ○ ○<br>(1) (2) (3) (4) (5) |
| 4 | The patient was brought to the operating room where he underwent coronary | ○ ○ ○ ○ ○<br>(1) (2) (3) (4) (5) |
| | artery bypass grafting to the coronary artery . | |
| | On postoperative day one , the patient was transferred to the floor in stable condition on neo- synephrine and propofol . | |
| | On postoperative day number five , the patient was discharged with home medications of home and was discharged home on postoperative day number five . | |
| | The patient was discharged to home on postoperative day number five , and afebrile was discharged on postoperative day number five , on postoperative day number five , the patient was discharged with home medications of visiting on postoperative day number five , and the patient was discharged home in good condition on the following follow-up . | |
| | The patient was discharged home on postoperative day number five afebrile , tolerating a regular diet , voiding all , and feeling well . | |
| 5 | The patient was admitted to the hospital and brought to the operating room on [deidt] for a coronary artery bypass graft . | |
| | Postoperatively on postoperative day number one , the patient 's hemodynamics status was improved and he was sent home with his home accordingly in stable condition . | |
| | He was discharged to home with vna for follow-up well as his primary care physician for his management well as hematocrit bun creatinine , and tsh aureus . | ○ ○ ○ ○ ○ <br> (1) (2) (3) (4) (5) |
| | The patient was discharged home on postoperative day number one and instructed to home with vna services on [deidt] . | |
| | The patient was to follow up with dr [deidt] in one to two weeks from the patient 's primary care physician in one week . | |
| 6 | The patient was admitted to the [deidt] surgical service . | |
| | The patient was taken to the operating room where he underwent coronary artery bypass grafting and four diagonal coronary artery bypass graft to the left anterior descending and left circumflex to the left anterior descending . | |
| | The patient tolerated the procedure well and was taken to the postoperative anesthesia care unit in stable condition . | |
| | The patient was transferred to the cardiothoracic intensive care unit for closer monitoring . | ○ ○ ○ ○ ○ <br> (1) (2) (3) (4) (5) |
| | The patient was discharged on postoperative day number five and had a temperature of 98.3 , and was subsequently monitored for a full postoperative course of v and was in the postoperative day . | |
| | The patient was discharged to home on postoperative day number five , instructed to home on postoperative day number five . | |
| Comment | | |

# Ch1. Appropriateness of entailment

## Question 3

| | Input |
|---|---|
| | The initial report stated that the indication for this examination was a diverticular abscess however , on review of the provided clinical information , the actual initial diagnosis was crohn 's disease complicated by multiple pelvic collections . |
| | The visualized abscesses are presumed to be related to the patient crohn 's current state . |
| | Subsequently , dilute intravenous contrast was injected via the pigtail catheter and the pelvis |
| | Small left-sided pleural effusion , this has decreased in size since the previous study . |
| | Both adrenal glands are unremarkable in appearance . non-contrast examination of the kidneys is unremarkable . |

| No. | Output | Rating 매우 이상함◀▶매우 적절함 |
|---|---|---|
| 1 | The patient was brought to the ct fluoroscopy suite and placed on the fluoroscopy table in prone position . The patient was brought to the angiography suite and placed prone on the imaging table , and the following which the guidewire was pulled back and the catheter was pulled back and pulled back to the level of the right renal vein . The catheter was then removed and a gentle hand injection of contrast through the sheath demonstrated the catheter to be located in the collection of the liver . The catheter was then removed and a hand injection of contrast through the catheter demonstrated a small amount of contrast trickling into the duodenum and proximal jejunum . The pigtail formed in the stomach and the duodenum and the pigtail formed in the duodenum and the pigtail formed in the stomach and the pigtail formed in the duodenum . | ○ (1) ○ (2) ○ (3) ○ (4) ○ (5) |
| 2 | The patient is status post cholecystectomy . The pancreas spleen , and right adrenal gland are normal in appearance . The left ureter is normal . The rectum , sigmoid colon , and bladder are normal in appearance . | ○ (1) ○ (2) ○ (3) ○ (4) ○ (5) |
| 3 | There is a cystic lesion seen in the body/tail of the pancreas , this has not changed significantly in the interval since the previous study . The small bowel is normal in caliber , no evidence of small bowel obstruction , no mesenteric fat stranding . There is a pigtail catheter in situ , low within the pelvis anteriorly within a loculated abscess . The anterior component , with the pigtail drain in situ has decreased in size slightly . Following discussion with the patient , an attempt was made to inject dilute | ○ (1) ○ (2) ○ (3) ○ (4) ○ (5) |

| | | | |
|---|---|---|---|
| | intravenous contrast via the pigtail catheter . | | |
| | The pigtail catheter was blocked at this point and had to be flushed with 10 ml of normal saline . | | |
| | This opacified a small cavity in the region of the pigtail however , most of the contrast material passed into the sigmoid colon and the rectum outlining a communicating tract between the diverticular abscess and the distal bowel . | | |
| | Specifically , there was no extravasation of contrast seen , no leakage of contrast around the catheter onto the skin . | | |
| | Therefore , it was determined by doctor. [deidt] not to up-size the catheter . | | |
| | There is communication between the anterior diverticular collection and the distal large bowel with contrast opacifying sigmoid and rectum . | | |
| 4 | The patient was prepped and draped in the usual sterile fashion . | | |
| | Approximately 15 cc of lidocaine was administered for local anesthesia . | | |
| | Under intermittent fluoroscopic guidance , a 20-ga needle was advanced into the collection . | | |
| | The catheter was attached to a drainage bag and secured to the skin . | ○ ○ ○ ○ ○ | |
| | The catheter was attached to a bag and was secured to the skin . | (1) (2) (3) (4) (5) | |
| | The patient tolerated the procedure well , and there were no immediate post-procedure complications . | | |
| | The catheter was secured to the skin and a sterile dressing was applied . | | |
| 5 | There is a small amount of fluid in the right paracolic gutter . | ○ ○ ○ ○ ○ | |
| | The right kidney is somewhat prominent but without focal lesion . | (1) (2) (3) (4) (5) | |
| | The right kidney is normal in appearance . | | |
| 6 | The spleen pancreas , adrenal glands kidneys stomach , and bowel loops are stable in appearance . | | |
| | The patient is status post coronary artery bypass grafting with patency of the right anterior descending and left circumflex coronary arteries . | ○ ○ ○ ○ ○ | |
| | The patient is status post median sternotomy and mitral valve replacement . | (1) (2) (3) (4) (5) | |
| | The patient tolerated the procedure well without immediate postprocedure complications . | | |
| | The patient received & fentanyl and versed throughout the procedure . | | |
| Comment | | | |

# Ch1. Appropriateness of entailment

## Question 4

| Input |
|---|
| [deidt] 9:58 am ercp biliary & pancreas by gi unit clip # [deidt] reason : please review ercp images from [deidt] admitting diagnosis : colon cancer\chemotherapy _ |
| 80 year old woman with necrotic mass in pancreas/porta hepatis and resulting biliary obstruction , jaundice reason for this examination : |
| One fluoroscopic spot view from ercp was submitted for review . |
| This image showss degenerative changes of the thoracolumbar spine and a calcified abdominal aorta . |

| No. | Output | Rating 매우 이상함 ◄►매우 적절함 |
|---|---|---|
| 1 | The patient tolerated the procedure well without immediate post- procedure complications . <br> The patient was placed prone on the angiographic table . <br> The patient was returned to the plastic ercp suite for further evaluation . | ○ ○ ○ ○ ○ <br> (1) (2) (3) (4) (5) |
| 2 | There is atelectasis of the lung bases . <br> Per ercp note , a large infiltrating mass was discovered in the wall of the duodenum and the ampulla could not be reached for evaluation of the biliary tree . <br> Further ercp images were not completed due to obstruction by a duodenal mass . | ○ ○ ○ ○ ○ <br> (1) (2) (3) (4) (5) |
| 3 | The patient was able to effectively sips liquids , with occasional liquids and the patient was able to cooperate . <br> During the procedure , the patient was able to stand the procedure for completion of the procedure . <br> Please see ercp note for further details . | ○ ○ ○ ○ ○ <br> (1) (2) (3) (4) (5) |
| 4 | Cannulation of the distal common bile duct and hepatic artery , with filling of the distal cbd and hepatic duct . <br> For further details , please refer to the gi endoscopy report on careweb . <br> Please refer to the gi ercp report on careweb for further details . | ○ ○ ○ ○ ○ <br> (1) (2) (3) (4) (5) |
| 5 | A left lower quadrant laser hip arthroplasty is unchanged in position . <br> The left lower quadrant was normal in width and signal intensity . <br> A plastic stent was placed within the left common iliac artery . <br> Please refer the report from the mra performed on the same day . | ○ ○ ○ ○ ○ <br> (1) (2) (3) (4) (5) |
| 6 | A plastic stent is seen within the common bile duct . <br> A plastic stent was placed in the right hepatic duct . <br> Please refer to ercp report in omr from [deidt] . | ○ ○ ○ ○ ○ <br> (1) (2) (3) (4) (5) |
| Comment | | |

# Ch1. Appropriateness of entailment

## Question 5

| | Input |
|---|---|
| | Final report history : 28-year-old man with pancreatitis complicated by ards , acute renal failure sirs , gi bleed , rule out pe . |
| | Technique : 2.5-mm contiguous axial images from the thoracic inlet through the adrenal glands with iv contrast were obtained . |
| | Coronal , sagittal and oblique mip images were included in this study . |
| | Comparison is made to a prior ct torso without iv contrast dated [deidt] . |

| No. | Output | Rating 매우 이상함◄►매우 적절함 |
|---|---|---|
| 1 | There is a small right pleural effusion with atelectasis at the left base . <br> The lungs are clear . <br> The heart and great vessels are normal in size . <br> There are no pathologically enlarged lymph nodes within the axilla mediastinum , or axilla . <br> There are no hilar , mediastinal or axillary lymph nodes according to ct size criteria of up to 15 mm in right and left lower paratracheal station . <br> There is no evidence of pulmonary embolism . <br> There are no bone findings of malignancy . | ○ ○ ○ ○ ○ <br> (1) (2) (3) (4) (5) |
| 2 | There is no evidence of pulmonary embolism . | ○ ○ ○ ○ ○ <br> (1) (2) (3) (4) (5) |
| 3 | There is no evidence of pulmonary embolus . <br> Interval worsening of the consolidation of the posterior aspects of the bilateral upper and lower lobes since the prior ct torso dated [deidt] . <br> In addition , there are nonspecific ground- glass opacities throughout both lungs . <br> Heart and great vessels appear normal . <br> Mediastinal and hilar lymph nodes measuring up to approximately 7 mm in short axis are visualized . <br> The tips of the right and left ij venous catheters are in the svc . <br> There is marked fatty infiltration of the liver . <br> Mild degenerative changes are seen in the lower thoracic spine . <br> Interval worsening of the consolidation of the dependent portions of the bilateral upper and lower lobes , which may represent atelectasis ; however , a superimposed infection can not be excluded . | ○ ○ ○ ○ ○ <br> (1) (2) (3) (4) (5) |
| 4 | The heart is normal in size . <br> The aorta is normal in caliber . <br> The lungs are clear . <br> The liver gallbladder spleen pancreas kidneys , and adrenal glands are normal . | ○ ○ ○ ○ ○ <br> (1) (2) (3) (4) (5) |
| | The left kidney is normal . | |
| | The bladder is normal in appearance . | |
| | The prostate is enlarged . | |
| 5 | There is a small amount of pericardial fluid , with a small amount of pericardial fluid . <br> The liver gallbladder spleen , and adrenal glands are normal in appearance . <br> The kidneys are symmetric in size and symmetric in signal intensity consistent with renal cysts . <br> The abdominal aorta and major tributaries are normal in caliber and contour . <br> The urinary bladder , distal ureters rectum , and sigmoid colon are normal in appearance . <br> There is a small ( grade ) ( over ) [deidt] 10:47 am <br> The left lower thoracic aortic aneurysm is not visualized ( the left transverse ) and the left renal artery origins are not well visualized ( off the images of the current study ) . | ○ ○ ○ ○ ○ <br> (1) (2) (3) (4) (5) |
| 6 | The heart and great vessels of the mediastinum are unremarkable . <br> The liver gallbladder spleen pancreas , adrenal glands kidneys stomach , and small bowel are within normal limits . <br> There is a small amount of free fluid in the pelvis . <br> The rectum , sigmoid colon uterus , and adnexa are unremarkable . <br> The osseous structures are unremarkable . | ○ ○ ○ ○ ○ <br> (1) (2) (3) (4) (5) |

Comment

# Ch1. Appropriateness of entailment

## Question 6

| | Input |
|---|---|
| | A catheter or pacing wire is seen in the ra and extending into the rv . |
| | Written informed consent was obtained from the patient . |
| | A tee was performed in the location listed above . |
| | The patient was monitored by a nurse |

| No. | Output | Rating 매우 이상함◄─►매우 적절함 |
|---|---|---|
| 1 | The patient appears to be in sinus rhythm .<br>Results were personally reviewed with the md caring for the patient .<br>The left atrium is mildly dilated .<br>No spontaneous echo contrast or thrombus is seen in the body of the left atrium/left atrial appendage or the body of the right atrium/right atrial appendage .<br>No atrial septal defect is seen by 2d or color doppler .<br>Overall left ventricular systolic function is normal ( lvef > 55 % ) .<br>Right ventricular chamber size and free wall motion are normal .<br>There are simple atheroma in the descending thoracic aorta . | ○ ○ ○ ○ ○<br>(1) (2) (3) (4) (5) |
| 2 | The patient was sedated for the tee .<br>Medications and dosages are listed above ( see test information section ) .<br>The patient appears to be in sinus rhythm .<br>Results were personally reviewed with the md caring for the patient .<br>The left atrium is mildly dilated .<br>No atrial septal defect is seen by 2d or color doppler .<br>There are simple atheroma in the aortic arch and descending thoracic aorta .<br>The aortic valve leaflets ( 3 ) appear structurally normal with good leaflet excursion and no aortic stenosis .<br>The mitral valve leaflets are structurally normal .<br>No mass or vegetation is seen on the mitral valve . | ○ ○ ○ ○ ○<br>(1) (2) (3) (4) (5) |
| 3 | The patient appears to be in sinus rhythm .<br>Results were personally reviewed with the md caring for the patient .<br>No spontaneous echo contrast or thrombus is seen in the body of the left atrium/left atrial appendage or the body of the right atrium/right atrial appendage .<br>No atrial septal defect is seen by 2d or color doppler .<br>There is mild symmetric left ventricular hypertrophy with normal cavity size and regional/global systolic function ( lvef > 55 % ) . | ○ ○ ○ ○ ○<br>(1) (2) (3) (4) (5) |
| | Right ventricular chamber size and free wall motion are normal . There are simple atheroma in the descending thoracic aorta . The aortic valve leaflets ( 3 ) appear structurally normal with good leaflet excursion and no aortic regurgitation . The mitral valve appears structurally normal with trivial mitral regurgitation . The left ventricular inflow pattern suggests impaired relaxation . | |
| 4 | The left atrium and right atrium are normal in cavity size . The right atrium is mildly dilated . The left ventricular cavity size is normal . Overall left ventricular systolic function is mildly depressed ( lvef= 45 % ) with a septal color doppler in the right atrium with systole color flow velocity of 133 mmhg ( increased color flow ) . The ascending aorta is mildly dilated and ascending valve leaflets ( 3 ) appear mildly thickened with mild aortic valve prolapse ( the aortic valve prosthesis ( 1.7cm 12.4 x ) with mild aortic prolapse ( fused the aortic valve leaflets ) 14 ) ( ) ( [deidt] ) ( - ( ) ) ( tee ) mitral regurgitation ( mac and non-coronary ( mitral valve directed below the aortic valve leaflets ( fused ) with leaflet flail ( with the aortic root ) and the ( of the ascending aorta ) with the aortic valve leaflets ( 3 ) and mildly thickened mitral valve prolapse ( the and ( ascending and descending thoracic aorta are normal in diameter and free of seen ( ascending and descending thoracic aortic leaflet ( | ○ ○ ○ ○ ○ (1) (2) (3) (4) (5) |
| 5 | The patient was sedated for the tee . Medications and dosages are listed above ( see test information section ) . The posterior pharynx was anesthetized with 2 % viscous lidocaine . The rhythm appears to be atrial fibrillation . Results were personally reviewed with the md caring for the patient . No spontaneous echo contrast is seen in the body of the left atrium or left atrial appendage . Overall left ventricular systolic function is normal ( lvef > 55 % ) . There are simple atheroma in the descending thoracic aorta . [deidt] was notified in person of the results immediately following the procedure . | ○ ○ ○ ○ ○ (1) (2) (3) (4) (5) |
| 6 | The patient was under general anesthesia throughout the procedure . The patient appears to be in sinus rhythm . Results were personally reviewed with the md caring for the patient . The left atrium is normal in size . No spontaneous echo contrast or thrombus is seen in the body of the left atrium/left atrial appendage or the body of the right atrium/right atrial appendage . | ○ ○ ○ ○ ○ (1) (2) (3) (4) (5) |

| | The left ventricular cavity size is normal . | |
| | Overall left ventricular systolic function is normal ( lvef > 55 % ) . | |
| | Right ventricular chamber size and free wall motion are normal . | |
| | There are simple atheroma in the descending thoracic aorta . | |
| | The mitral valve leaflets are mildly to moderately thickened and calcified . | |
| | No mass or vegetation is seen on the mitral valve . | |

Comment

# Ch1. Appropriateness of entailment

## Question 7

| | Input |
|---|---|
| | Following intravenous injection of autologous red blood cells labeled with tc- [deidt] m , blood flow and dynamic images of the abdomen for 90 minutes were obtained . |
| | The patient was again imaged dynamically for 20 one minute images at 9 hours after the original injection . |
| | Left lateral views of the pelvis were also obtained . |
| | Blood flow images show tortuosity of the aorta , but no bleeding site . |
| | Dynamic images show no active bleeding site , however left lateral images show activity in the region of the rectum which appears to change its configuration on each lateral view , suggesting a rectal bleeding site . |

| No. | Output | Rating<br>매우 이상함◄►매우 적절함 |
|---|---|---|
| 1 | Correlation with the initial study shows no evidence of active bleeding .<br>[deidt] , m.d . approved : [deidt] 4:28 pm radline [deidt] ; a radiology consult service .<br>To hear preliminary results , prior to transcription , call the radiology listen line [deidt] . | ○ ○ ○ ○ ○<br>(1) (2) (3) (4) (5) |
| 2 | Findings discussed with doctor. [deidt] .<br>[deidt] , m.d . approved : mon [deidt] 3:13 pm<br>To hear preliminary results , prior to transcription , call the radiology listen line [deidt] . | ○ ○ ○ ○ ○<br>(1) (2) (3) (4) (5) |
| 3 | There is no evidence of active extravasation of the blood product .<br>The findings correspond to a functional perfusion defect in the distal common bile duct , consistent with a known bleed in this region .<br>[deidt] , m.d . approved : [deidt] 4:09 pm radline [deidt] ; a radiology consult service .<br>To hear preliminary results , prior to transcription , call the radiology listen line [deidt] . | ○ ○ ○ ○ ○<br>(1) (2) (3) (4) (5) |
| 4 | The patient was admitted to the hospital with platelet count of 14,000 to the minute of admission 's day number of the day of the day .<br>[deidt] , m.d , [deidt] , m.d . approved : [deidt] 3:12 pm radline [deidt] ; a radiology consult service .<br>To hear preliminary results , prior to transcription , call the radiology listen line , [deidt] , call , [deidt]<br>[deidt] , m.d [deidt] was notified in the findings of dictation ( over [deidt] | ○ ○ ○ ○ ○<br>(1) (2) (3) (4) (5) |
| 5 | The [deidt] study m.d was obtained with tl-201 octreotide , and post exercise .<br>The findings correspond to the findings of the gi tract .<br>The kidneys and urinary bladder are visualized , the normal route of tracer | ○ ○ ○ ○ ○<br>(1) (2) (3) (4) (5) |
|   | excretion . [deidt] , m.d . approved : mon [deidt] To hear preliminary results , prior to transcription , call the radiology listen line [deidt] . |   |
| 6 | At [deidt] , m.d . approved : mon [deidt] 3:10 pm To hear preliminary results , prior to transcription , call the radiology listen line [deidt] . [deidt] , m.d . approved : [deidt] 12:07 am radline [deidt] ; a radiology consult service . | ○ ○ ○ ○ ○ (1) (2) (3) (4) (5) |
| Comment | | |

# Ch1. Appropriateness of entailment

## Question 8

| | Input |
|---|---|
| | Rule out progression or new lesions |
| | Findings : again noted are multiple enhancing lesions in both cerebral hemispheres involving both frontal lobes , left posterior temporal lobe and both inferior cerebellar regions . |
| | The enhancing lesions have slightly decreased compared to the prior study . |
| | The edema surrounding both frontal convexity lesions has slightly increased which could be related to therapy . |
| | Otherwise the edema noted at other lesions in both frontal lobes and the left temporal lobe has slightly decreased . |

| No. | Output | Rating 매우 이상함 ◄──► 매우 적절함 |
|---|---|---|
| 1 | There is no mass effect or shift of normally seen midline structures . The ventricles and sulci are normal in caliber and configuration . The visualized paranasal sinuses and osseous structures are unremarkable . | ○(1) ○(2) ○(3) ○(4) ○(5) |
| 2 | Multiple brain metastatic lesions as described above have decreased in size since the previous mri examination of [deidt] . The right inferior frontal lobe lesion measured 1.1 cubic centimeter on the previous study and now measures approximately 0.55 cubic centimeter . The right frontal convexity lesion measured 950 cubic millimeter on the previous study and measures 595 cubic millimeter on the current study . Similarly the other lesions also decreased in volume which include the right cerebellar lesion which measured 44.3 cubic millimeter , right frontal convexity lesion which measures 595 cubic millimeter . | ○(1) ○(2) ○(3) ○(4) ○(5) |
| 3 | There is also a new area of flair hyperintensity in the right frontal lobe . The ventricles are stable in size . The left frontal ventriculostomy catheter is in unchanged position . The visualized paranasal sinuses are clear . | ○(1) ○(2) ○(3) ○(4) ○(5) |
| 4 | The right frontal lobe lesion is unchanged in size and appearance . | ○(1) ○(2) ○(3) ○(4) ○(5) |
| 5 | There is no evidence of hydrocephalus . The ventricles and sulci are normal in size and configuration . The visualized paranasal sinuses and mastoid air cells are clear . | ○(1) ○(2) ○(3) ○(4) ○(5) |
| 6 | There is a small amount of right cerebellar infarct . The right frontal and temporal bone : there is a small ( over ) [deidt] 3:12 pm The right internal jugular vein is not clearly identifiable on the study . The right internal carotid artery is patent and demonstrates mild to moderate mitral regurgitation . The left internal carotid and the posterior cerebral arteries are patent with normal respirophasic variation and the caliber of the right vertebral artery is diminutive | ○(1) ○(2) ○(3) ○(4) ○(5) |

| | |
|---|---|
| and the caliber of the right vertebral artery is diminutive and the caliber of the right vertebral artery is normal in appearance and the remainder of the brain parenchyma is preserved .<br>The right frontal and right parietal and occipital leptomeningeal enhancement are unchanged in appearance .<br>The mastoid air cells and paranasal sinuses are clear .<br>The volumetric study of the brain demonstrates no significant change in the pattern of enhancement in the soft tissues . | |

Comment

# Ch1. Appropriateness of entailment

## Question 9

| Input |
|---|
| This study was compared to the report of the prior study ( images not available ) of [deidt] . |
| regional lv wall motion abnormalities include : basal inferior - hypo ; mid inferior - hypo ; basal inferolateral - hypo ; mid inferolateral - hypo ; right ventricle : normal rv chamber size and free wall motion . |
| The left ventricular cavity size is top normal/borderline dilated . |
| Overall left ventricular systolic function is mildly depressed . |
| Resting regional wall motion abnormalities include inferior and inferolateral hypokinesis with mild hypokinesis elsewhere . |

| No. | Output | Rating<br>매우 이상함◀▶매우 적절함 |
|---|---|---|
| 1 | Right ventricular chamber size and free wall motion are normal .<br>The ascending aorta is moderately dilated .<br>The aortic arch is mildly dilated .<br>Compared with the report of the prior study ( images unavailable for review ) of [deidt] , left ventricular function may be similar or slightly worse but prior study not available for direct comparison .<br>Mitral regurgitation is now more prominent . | ○ ○ ○ ○ ○<br>(1) (2) (3) (4) (5) |
| 2 | Right ventricular chamber size and free wall motion are normal .<br>The diameters of aorta at the sinus , ascending and arch levels are normal .<br>The aortic valve leaflets ( 3 ) appear structurally normal with good leaflet excursion and no aortic regurgitation .<br>The mitral valve appears structurally normal with trivial mitral regurgitation .<br>Compared with the report of the prior study ( images unavailable for review ) of [deidt] , lv function is not as dynamic on the current study . | ○ ○ ○ ○ ○<br>(1) (2) (3) (4) (5) |
| 3 | The left ventricular cavity size is normal .<br>There is moderate regional left ventricular systolic dysfunction with hypokinesis of the inferior septum and anterior free wall hypokinesis . Compared with the prior study ( images reviewed ) of [deidt] , the findings are similar . | ○ ○ ○ ○ ○<br>(1) (2) (3) (4) (5) |
| 4 | Right ventricular chamber size and free wall motion are normal .<br>The aortic root is mildly dilated at the sinus level .<br>The ascending aorta is mildly dilated .<br>The aortic valve leaflets ( 3 ) appear structurally normal with good leaflet excursion and no aortic regurgitation .<br>The mitral valve leaflets are structurally normal .<br>The estimated pulmonary artery systolic pressure is normal .<br>Compared with the report of the prior study ( images unavailable for review ) of | ○ ○ ○ ○ ○<br>(1) (2) (3) (4) (5) |
| | [deidt] , the left ventricle is more hypertrophied . <br> The severity of mitral regurgitation is reduced . | |
| 5 | Right ventricular chamber size and free wall motion are normal . <br> The aortic root is mildly dilated at the sinus level . <br> The ascending aorta is mildly dilated . <br> The aortic valve leaflets ( 3 ) appear structurally normal with good leaflet excursion and no aortic stenosis or aortic regurgitation . <br> The mitral valve appears structurally normal with trivial mitral regurgitation . <br> Compared with the report of the prior study ( images unavailable for review ) of [deidt] , left ventricular systolic function has improved . | ○ ○ ○ ○ ○ <br> (1) (2) (3) (4) (5) |
| 6 | Right ventricular chamber size is normal . <br> The aortic root is mildly dilated at the sinus level . <br> The ascending aorta is mildly dilated . <br> The aortic valve leaflets ( 3 ) appear structurally normal with good leaflet excursion and no aortic regurgitation . <br> The mitral valve appears structurally normal with trivial mitral regurgitation . <br> Compared with the report of the prior study ( images unavailable for review ) of [deidt] , the findings are similar . | ○ ○ ○ ○ ○ <br> (1) (2) (3) (4) (5) |
| Comment | | |

# Ch1. Appropriateness of entailment

## Question 10

| | Input |
|---|---|
| | Admitting diagnosis : low white blood count , elevated lst _ |
| | Assess blood flows , presence of fluid/abscess _ |
| | The small heterogeneous area within the right lobe on ct was not visualized despite attempting to correlate directly . |
| | The main portal vein left , and right portal vein branches appear unremarkable on color and duplex imaging . |
| | The waveforms involving the main hepatic artery well as the left and right hepatic arteries appear unremarkable , i.e . brisk upstroke and antegrade flow during diastole . |

| No. | Output | Rating 매우 이상함 ◄──► 매우 적절함 |
|---|---|---|
| 1 | The left hepatic artery was not identified . | ○(1) ○(2) ○(3) ○(4) ○(5) |
| 2 | The main right , and left hepatic arteries are patent with normal waveforms . | ○(1) ○(2) ○(3) ○(4) ○(5) |
| 3 | The main portal vein is patent with hepatopetal flow and a velocity of 55 cm/sec . The main hepatic artery is patent with appropriate waveforms . The left hepatic artery is patent and demonstrates normal waveform . The left hepatic artery is not visualized , but is diminutive . | ○(1) ○(2) ○(3) ○(4) ○(5) |
| 4 | The main portal vein is patent with antegrade flow . The main hepatic artery and the segmental arteries of the upper , mid lower and upper portions of the left kidney demonstrate normal systolic upstrokes and resistive indices ranging from 0.56 to 0.69 . The right hepatic artery is not clearly visualized and is normal in appearance . The right hepatic artery is not well visualized and is likely secondary to technical impediments . The right hepatic artery is not demonstrated , which may be due to technical reasons . The right hepatic artery is not visualized and may be technical factors . The findings were discussed with doctor. [deidt] at the time of dictation . | ○(1) ○(2) ○(3) ○(4) ○(5) |
| 5 | The right main stem and right hepatic veins are patent with the hepatic veins and patent branch of the left portal vein . The right hepatic artery is patent and demonstrates normal systolic function with a resistive index of 0.56 ( 0.64 and 0.4 cm ) . The right hepatic vein and the left portal vein are patent with flow in the appropriate direction . The right and left portal veins are patent with hepatopetal flow and a velocity of 28 cm per second . | ○(1) ○(2) ○(3) ○(4) ○(5) |

| 6 | Elevated velocity within the main hepatic vein is most likely due to ( tortuosity as indicated on ct ) . | ○ ○ ○ ○ ○ |
| | The hepatic veins all appear patent with appropriate direction of flow . | (1) (2) (3) (4) (5) |
| | There is a question of filling defect within the distal common bile duct which shows no shadowing . | |
| | The duct proximal to this area does not appear to be dilated . | |
| | This is likely artifactual . | |

Comment

# [Chapter 2. Discrimination]

<Instruction>

Questions in this chapter will ask you to identify text written by human among a set of texts. Please opt human's writing. If you can identify human's writing, please provide us the reason you can find the answer.

☆ [deidt] is a label replacing de-identification texts.

# Ch2. Discrimination

## Question 1

| 1 | Left lower lobe retrocardiac atelectasis is unchanged . |
|---|---|
| 2 | The ascending aorta is mildly dilated . <br><br> The aortic valve leaflets ( 3 ) appear structurally normal with good leaflet excursion and no aortic regurgitation . <br><br> The tricuspid valve appears structurally normal with trivial tricuspid regurgitation . <br><br> The left ventricular cavity size is normal . <br><br> Overall left ventricular systolic function is mildly depressed ( lvef= 30-35 % ) with a focal area of increased in the right apex ( apical segment ) . <br><br> Right ventricular chamber size and free wall motion are normal . <br><br> There are simple atheroma in the descending thoracic aorta . |
| 3 | There is mild symmetric left ventricular hypertrophy with normal cavity size and regional/global systolic function ( lvef > 55 % ) . <br><br> Right ventricular chamber size and free wall motion are normal . <br><br> The left ventricular cavity size is normal . <br><br> There is moderate to severe global left ventricular free wall hypokinesis . <br><br> There is moderate to severe pulmonary artery systolic hypertension . |

**Please identify human's writing among above options.**

☐ no.1      ☐ no.2

☐ no.3      ☐ none of them

**Was it confused to choose the answer?**

☐ easy      ☐ normal      ☐ confused

# Ch2. Discrimination

## Question 2

| | |
|---|---|
| 1 | There is no evidence of pneumothorax . |
| | The left subclavian central venous catheter tip is at the cavoatrial junction . |
| | The left ventricular cavity size is normal . |
| | There is moderate to severe hypokinesis of the left ventricle . |
| | The right ventricular cavity is mildly dilated with borderline normal free wall function . |
| | The aortic root is mildly dilated at the sinus level . |
| | The left ventricular inflow pattern suggests a restrictive filling abnormality , with elevated left atrial pressures . |
| 2 | The ascending aorta is mildly dilated at its level ( 3 cm ) with the tip regurgitant to the aortic valve . |
| | The mitral valve appears structurally normal with trivial mitral regurgitation . |
| | The left ventricular cavity size is normal . |
| | Overall left ventricular systolic function is mildly depressed ( lvef= 30-35 % ) with the apex ( `` notch ) and the right ventricular cavity size is increased ( tr ) to the right and left sided free intraperitoneal c/w a pericardial effusion . |
| | The diameters of aorta at the sinus , ascending and arch levels are normal ( the patient ) and the aortic valve leaflets are mildly thickened with mild mitral regurgitation ( the aortic valve prolapse ) to left ) prolapse of the and right raphe ( mitral valve prolapse ) with ( thickened ) tr echo : |
| | There is a mitral valve annuloplasty in the expected mitral position with a thickened rheumatic ring in systole c/w mitral regurgitation . reason : please evaluate for reposition of r admitting diagnosis : cellulitis |
| 3 | The lungs are clear . |
| | The heart size is normal . |
| | There is no evidence of pneumothorax . |

**Please identify human's writing among above options.**

☐ no.1    ☐ no.2
☐ no.3    ☐ none of them

**Was it confused to choose the answer?**

☐ easy    ☐ normal    ☐ confused

# Ch2. Discrimination

## Question 3

| | |
|---|---|
| 1 | There is no significant change in the appearance of the left frontal subdural hematoma . |
| | There is no evidence of hydrocephalus . |
| | The left lateral ventricle is not clearly seen , but the left lateral ventricle is not clearly seen . |
| 2 | The ventricles and sulci are normal in size and configuration . |
| | There is no evidence of hemorrhage edema masses , mass effect , or infarction . |
| | The osseous and soft tissue structures are unremarkable . |
| 3 | Contiguous axial images through the brain were obtained without intravenous contrast . |
| | Findings : tiny foci of subarachnoid hemorrhage within the left parietal lobe are largely resolved from the prior study . |
| | A subdural hematoma layering over the left parieto-occipital convexity is again identified , measuring 4 mm in thickness . |
| | Subdural hematoma layering along the left tentorium and falx cerebri appears unchanged . |
| | No new foci of hemorrhage are seen . |
| | There is no shift of normally midline structures hydrocephalus , mass or vascular territorial infarct . |
| | Multiple skin staples overlie the right frontal convexity . |
| | No interval change in subdural hematoma along the falx cerebri , left tentorium and extending along the left parieto-occipital convexity . |
| | No mass effect or new areas of hemorrhage seen . |

**Please identify human's writing among above options.**

☐ no.1   ☐ no.2
☐ no.3   ☐ none of them

**Was it confused to choose the answer?**

☐ easy   ☐ normal   ☐ confused

# Ch2. Discrimination

## Question 4

| | |
|---|---|
| 1 | Et tube is in standard position . <br> Ng tube passes below the diaphragm and out of view . |
| 2 | The ascending aorta is mildly dilated . <br> The aortic valve leaflets ( 3 ) appear structurally normal with good leaflet excursion and no aortic regurgitation . <br> The mitral valve appears structurally normal with trivial mitral regurgitation . <br> The left ventricular cavity size is normal . <br> Overall left ventricular systolic function is mildly depressed ( lvef= 30-35 % ) with a focal area of increased transmitral fat in the right ventricle . <br> The diameters of aorta at the sinus , ascending and arch levels are normal . |
| 3 | There is no evidence of pneumothorax . |

**Please identify human's writing among above options.**

☐ no.1　　　☐ no.2

☐ no.3　　　☐ none of them

**Was it confused to choose the answer?**

☐ easy　　　☐ normal　　　☐ confused

# Ch2. Discrimination

## Question 5

| | |
|---|---|
| 1 | Although the amount of hematoma is still large , it has significantly decreased compared to the prior study . Decreased perinephric inflammation and fluid is also seen compared to the prior study . The patient 's known right lower pole renal mass is not well seen due to limited bolus timing well as compression of the renal parenchyma by the hematoma . Multiple stable left renal lesions are seen , as previously characterized by mri to be simple cysts . Within the anterior mid-abdominal mesentery , there is a region of stranding with multiple mildly prominent lymph nodes . The solid and hollow organs and the remainder of the mesentery are within normal limits . There are degenerative changes of the spine . Note is made of an ax-fem bypass within the left lateral subcutaneous tissues . Large but decreased right renal subcapsular hematoma and perinephric fluid , with limited visualization of the patient 's known right lower pole renal mass , as described . |
| 2 | There is a nasojejunal tube in place . There is a percutaneous drain in the right hepatic lobe . There is a percutaneous biliary catheter in place , with the tip of the pigtail catheter in the right hepatic lobe . There is a small amount of free fluid in the pelvis . There is a foley catheter within the bladder . |
| 3 | There is no evidence of bowel obstruction . The spleen pancreas , and adrenal glands are normal in appearance . The kidneys are small and contain bilaterally . There is no free fluid or free air in the abdomen . |

**Please identify human's writing among above options.**

☐ no.1    ☐ no.2
☐ no.3    ☐ none of them

**Was it confused to choose the answer?**

☐ easy    ☐ normal    ☐ confused

# Ch2. Discrimination

## Question 6

| | |
|---|---|
| 1 | The mediastinal and hilar contours are normal . <br> There is no evidence of chf . <br> The osseous structures are unremarkable . |
| 2 | The aorta is tortuous and calcified . <br> The right pleural effusion is slightly increased in size . |
| 3 | The pulmonary vessels are within normal limits and there is no evidence to indicate cardiac failure . <br> A drainage tube is noted overlying the upper abdomen . <br> The left picc line remains well positioned . |

**Please identify human's writing among above options.**

☐ no.1   ☐ no.2

☐ no.3   ☐ none of them

**Was it confused to choose the answer?**

☐ easy   ☐ normal   ☐ confused

# Ch2. Discrimination

## Question 7

| 1 | There is a 13-mm barium tablet freely into the stomach . |
| --- | --- |
| | There is a 13-mm stretch of barium into the stomach . |
| | There is a small amount of intraperitoneal free fluid . |
| 2 | There is no hiatal hernia seen and no gastroesophageal reflux demonstrated despite provocative maneuvers . |
| | There is no significant retention in the valleculae or piriform sinuses . |
| | There is a very small amount of penetration into the vestibule with no aspiration into the airway . |
| | A 13-mm barium tablet pauses transiently at the ge junction and ultimately passes freely into the stomach . |
| 3 | There is a small amount of residual barium contrast material in the oropharynx . |
| | There is a small amount of barium material in the stomach . |

**Please identify human's writing among above options.**

☐ no.1    ☐ no.2

☐ no.3    ☐ none of them

**Was it confused to choose the answer?**

☐ easy    ☐ normal    ☐ confused

# Ch2. Discrimination

## Question 8

| 1 | The right groin was prepped and draped in the standard sterile fashion .<br>Under fluoroscopic guidance , a 21-gauge chiba needle was advanced into the right groin .<br>A 5 french [deidt] catheter was advanced into the left common femoral artery and connected to continuous saline infusion .<br>The catheter was then exchanged for a 0.035 angled glidewire , which was used to catheterize the left internal carotid artery .<br>The catheter was then exchanged for a 5 french vascular sheath which was placed in the right common femoral artery .<br>The catheter was then sutured to the skin and connected to a continuous saline flush .<br>The catheter was secured to the skin with 0 silk suture and sterile dressing applied . |
|---|---|
| 2 | The catheter was then removed and the sheath was then removed and hemostasis was achieved .<br>The catheter was then removed over the wire , and a 4 french sheath was placed in the superior mesenteric artery .<br>The sheath was then removed and pressure was held on the right groin until hemostasis was achieved .<br>The catheter was then flushed and capped in the post- postop fashion .<br>The patient tolerated the procedure well .<br>The catheter was placed to external bag drainage . |
| 3 | The catheter was then removed over the wire and a 4-french bright catheter was advanced over the wire into the superior mesenteric artery .<br>The catheter was then removed over the wire and a 4-french c2 cobra catheter was advanced over the glidewire into the superior mesenteric artery .<br>The microcatheter was then removed and the [deidt] catheter was advanced through the sheath into the superior mesenteric artery and the arteriogram was then obtained .<br>The microcatheter was then removed and the catheter was advanced through the sheath into the right common iliac artery angiogram . |

**Please identify human's writing among above options.**

☐ no.1  ☐ no.2
☐ no.3  ☐ none of them

**Was it confused to choose the answer?**

☐ easy  ☐ normal  ☐ confused

# Ch2. Discrimination

## Question 9

| 1 | A 2-cm right shift of the midline structures is stable . |
| --- | --- |
| | Marked mass effect on the corpus callosum and lateral ventricles is noted . |
| | The visualized soft tissue structures appear unremarkable . |
| | In comparison to [deidt] exam , there is no significant interval change in size and appearance of the left frontal extra-axial mass lesion . |
| | There is no evidence of interval hemorrhage or worsening edema . |
| 2 | There is a small amount of hyperdense material layering within the occipital horns of the lateral ventricles . |
| | There is a small amount of blood layering in the occipital [deidt] of the left lateral ventricle . |
| | There is a small amount of intraventricular blood in the right frontal lobe . |
| | The visualized paranasal sinuses and mastoid air cells are clear . |
| 3 | There is no evidence of acute hemorrhage . |

**Please identify human's writing among above options.**

☐ no.1  ☐ no.2
☐ no.3  ☐ none of them

**Was it confused to choose the answer?**

☐ easy  ☐ normal  ☐ confused

# Ch2. Discrimination

## Question 10

| 1 | The right and left kidneys are well expanded and display no appreciable abnormality . |
|   | The left kidney is normal in size and appearance . |
| 2 | The right abi is 0.84 . |
|   | On the left there is a biphasic pt waveform and a monophasic dp waveform . |
|   | Pulse volume recordings are dampened in the calf bilaterally . |
|   | This may be due to severe leg swelling . |
| 3 | Abis is consistent with acute arterial disease . |
|   | Pulse volume recordings showed a drop off from the right forearm to the popliteal artery . |
|   | On the left , peak systolic velocities are 156 186 , and 215 cm/sec in the proximal , mid and distal femoral respectively , and at the mid to distal anastomosis . |
|   | The ankle-brachial index is 1.06 on the right and 1.46 on the left . |

**Please identify human's writing among above options.**

☐ no.1   ☐ no.2
☐ no.3   ☐ none of them

**Was it confused to choose the answer?**

☐ easy   ☐ normal   ☐ confused

-Fin-